%% file: main.tex
\documentclass[preprint,12pt]{elsarticle}

\usepackage{amssymb}
\usepackage{amsthm}
\usepackage{amsmath}
\usepackage{tikz}
\usepackage{algorithmic}
\usepackage{algorithm}
\usepackage{subcaption}
\PassOptionsToPackage{hyphens}{url}\usepackage{hyperref}
\usepackage{xurl}

\usepackage{breqn}

\usepackage{array}
\newcommand{\PreserveBackslash}[1]{\let\temp=\\#1\let\\=\temp}
\newcolumntype{C}[1]{>{\PreserveBackslash\centering}p{#1}}
\newcolumntype{M}[1]{>{\PreserveBackslash\centering}m{#1}}
\newcolumntype{R}[1]{>{\PreserveBackslash\raggedleft}p{#1}}
\newcolumntype{L}[1]{>{\PreserveBackslash\raggedright}p{#1}}

\journal{Information Sciences}

\begin{document}

\begin{frontmatter}

%% Title, authors and addresses

%% use the tnoteref command within \title for footnotes;
%% use the tnotetext command for theassociated footnote;
%% use the fnref command within \author or \address for footnotes;
%% use the fntext command for theassociated footnote;
%% use the corref command within \author for corresponding author footnotes;
%% use the cortext command for theassociated footnote;
%% use the ead command for the email address,
%% and the form \ead[url] for the home page:
%% \title{Title\tnoteref{label1}}
%% \tnotetext[label1]{}
%% \author{Name\corref{cor1}\fnref{label2}}
%% \ead{email address}
%% \ead[url]{home page}
%% \fntext[label2]{}
%% \cortext[cor1]{}
%% \affiliation{organization={},
%%             addressline={},
%%             city={},
%%             postcode={},
%%             state={},
%%             country={}}
%% \fntext[label3]{}

\title{Effect of Choosing Loss Function when Using T-batching for Representation Learning on Dynamic Networks}

%% use optional labels to link authors explicitly to addresses:
%% \author[label1,label2]{}
%% \affiliation[label1]{organization={},
%%             addressline={},
%%             city={},
%%             postcode={},
%%             state={},
%%             country={}}
%%
%% \affiliation[label2]{organization={},
%%             addressline={},
%%             city={},
%%             postcode={},
%%             state={},
%%             country={}}

\author[inst1]{Erfan~Loghmani}
\author[inst1]{MohammadAmin~Fazli}

\affiliation[inst1]{organization={Department
of Computer Engineering, Sharif University of Technology},%Department and Organization
            addressline={Azadi St.}, 
            city={Tehran},
            postcode={1458889694}, 
            state={Tehran},
            country={Iran}}
% \author[inst1]{Author Three}

%\affiliation[inst2]{organization={Department Two},%Department and Organization
%            addressline={Address Two}, 
%            city={City Two},
%            postcode={22222}, 
%            state={State Two},
%            country={Country Two}}

\begin{abstract}
%% Text of abstract

Representation learning methods have revolutionized machine learning on networks by converting discrete network structures into continuous domains. However, dynamic networks that evolve over time pose new challenges. To address this, dynamic representation learning methods have gained attention, offering benefits like reduced learning time and improved accuracy by utilizing temporal information.

T-batching is a valuable technique for training dynamic network models that reduces training time while preserving vital conditions for accurate modeling. However, we have identified a limitation in the training loss function used with t-batching. Through mathematical analysis, we propose two alternative loss functions that overcome these issues, resulting in enhanced training performance.

We extensively evaluate the proposed loss functions on synthetic and real-world dynamic networks. The results consistently demonstrate superior performance compared to the original loss function. Notably, in a real-world network characterized by diverse user interaction histories, the proposed loss functions achieved more than 26.9\% enhancement in Mean Reciprocal Rank (MRR) and more than 11.8\% improvement in Recall@10. These findings underscore the efficacy of the proposed loss functions in dynamic network modeling.

% Representation learning methods on networks have enabled using machine learning methods on networks' discrete structures by transferring their information to a continuous domain. As networks' structures are not always static and may evolve through time, dynamic representation learning methods have recently gained attention. These methods could significantly reduce the learning time by refreshing the model as the changes occur, and improve accuracy by using the temporal information of interactions.

% T-batching is a technique used for training representation learning models on dynamic networks. This technique reduces training time using the batching idea while preserving conditions that are vital in dynamic network modeling. This research indicates a problem with the training loss function used with t-batching in a previous study. By mathematically analyzing the loss function, we show its downsides and suggest two other loss functions that do not suffer from the original function's problems. 
% Then, we study the effect of loss function on the model's accuracy and optimization by designing several experiments on real and generated dynamic networks. These experiments show that alternative loss functions could improve some evaluation metrics on the edge prediction task from $1.8\%$ to $27.4\%$ depending on some of the network's structural properties.
\end{abstract}

%%Graphical abstract
%\begin{graphicalabstract}
%\includegraphics{grabs}
%\end{graphicalabstract}

%%Research highlights
\begin{highlights}
    \item Indicated a problem in training dynamic network representation learning with t-batching and theoretically analyzed the impact of the problem on model performance.
    \item Proposed two alternative loss functions that address the bias caused by varying batch sizes; Evaluated their effectiveness through comprehensive experiments on synthetic and real-world dynamic networks, showcasing significant improvements compared to the original loss function.
    \item Investigated the characteristics of dynamic networks that impact loss function performance during the training process, shedding light on the factors that influence model accuracy and optimization.
    
    %\item Introduce two loss functions that don’t suffer from the original loss function problems and show their effectiveness 
    %\item Investigate the characteristics of a dynamic network and training process that affect loss function performance
\end{highlights}

\begin{keyword}
%% keywords here, in the form: keyword \sep keyword
Dynamic Networks \sep Representation Learning \sep Deep Learning \sep Loss Function
%% PACS codes here, in the form: \PACS code \sep code
%\PACS 0000 \sep 1111
%% MSC codes here, in the form: \MSC code \sep code
%% or \MSC[2008] code \sep code (2000 is the default)
%\MSC 0000 \sep 1111
\end{keyword}

\end{frontmatter}

%% \linenumbers

%% main text
\input{Introduction}

\input{Setup}
\input{Theory}
\input{Results}

\section{Conclusion}
% This paper investigated the loss function used during the t-batching mechanism in dynamic network representation learning methods. By analyzing the t-batching in detail, we found a problem in the loss function used during this process. The issue arises because of the loss function’s sensitivity to batch sizes and unequal batch sizes during t-batching. We introduced two loss functions that resolved this problem and, by theoretical and practical evidence, showed how the primary loss function could be problematic. We showed that proposed loss functions could enhance the model’s accuracy and help the optimization process.

In this paper, we investigated the impact of the loss function in JODIE, a model for dynamic network embeddings that utilizes t-batching to process interactions in the order they occur. We identified a problem with the original loss function that could lead to biased predictions, particularly in scenarios where the batch sizes vary significantly. To address this issue, we proposed two alternative loss functions, $loss_{item-sum}$ and $loss_{full-sum}$, which modify the regularization terms to ensure smoother updates of node embeddings during training.

Through extensive experiments on both synthetic and real-world dynamic networks, we demonstrated the effectiveness of the proposed loss functions in mitigating the bias problem. Our analyses revealed that the original loss function could lead to incorrect predictions in certain cases, whereas the $loss_{item-sum}$ and $loss_{full-sum}$ functions consistently outperformed it in terms of prediction accuracy.

Additionally, we explored the performance of the loss functions on various network characteristics, such as the diversity of user interaction histories and batch size distributions. Our results indicated that the introduced loss functions consistently yielded improvements across different datasets, and the improvement is more pronounced when the interaction histories are more diverse and batch sizes have higher variation.
%distribution is more  affirming their robustness and applicability in various scenarios.

%Furthermore, we compared the loss functions' behaviors on synthetic networks with different characteristics, ranging from purely time-independent random structures to more complex deterministic and Markov models. The experimental findings reinforced the effectiveness of the proposed loss functions, showcasing their ability to handle diverse network generation processes.

In conclusion, our study highlights the importance of carefully designing loss functions for dynamic network embedding models, especially when processing interactions using t-batching. The proposed $loss_{item-sum}$ and $loss_{full-sum}$ functions address the bias issue and lead to improved prediction accuracy in various network scenarios. By considering the characteristics of the networks and the impact of t-batching on batch size distributions, we can better understand the behavior of loss functions in dynamic network embedding tasks. Our research contributes to the advancement of dynamic network embedding methodologies, enabling more accurate and robust modeling of evolving networks across different application domains.

\section*{Acknowledgment}

The authors express their gratitude to Myket corporation for generously providing the anonymous interaction data used in this research.
We are grateful to Dr. Vahid Rahimian for agreeing to collaborate with us and for his consistent follow-up throughout the process. We would also like to express our appreciation to Ms. Zahra Eskandari for her proactive assistance in gathering and organizing the data.

\begin{appendix}
\section{Supplementary material}
The code for this research project, including the implementation of alternative loss functions, is accessible on GitHub at the following link: 
%\href{https://github.com/erfanloghmani/effect-of-loss-function-tbatching}{github.com/erfanloghmani/effect-of-loss-function-tbatching}
\url{https://github.com/erfanloghmani/effect-of-loss-function-tbatching}
. Additionally, the Myket dataset, containing application install interactions for users in the Myket Android application market, is available on GitHub at \url{https://github.com/erfanloghmani/myket-android-application-market-dataset}. Researchers and interested parties can access both the code and dataset for further exploration and reproduction of the experiments conducted in this study.
\end{appendix}

%We should thank Dr. Vahid Rahimian for accepting our collaboration request and follow-ups along the path. Also, from Ms. Zahra Eskandari for her active support in cleaning \& collecting data.

%The authors would like to thank Myket corporation for its help in providing anonymous interaction data. We should thank Dr. Vahid Rahimian for accepting our collaboration request and follow-ups along the path. Also, from Ms. Zahra Eskandari for her active support in cleaning \& collecting data.

%% The Appendices part is started with the command \appendix;
%% appendix sections are then done as normal sections
% \appendix

%\section{Sample Appendix Section}
%\label{sec:sample:appendix}
%Lorem ipsum dolor sit amet, consectetur adipiscing elit, sed do eiusmod tempor section \ref{sec:sample1} incididunt ut labore et dolore magna aliqua. Ut enim ad minim veniam, quis nostrud exercitation ullamco laboris nisi ut aliquip ex ea commodo consequat. Duis aute irure dolor in reprehenderit in voluptate velit esse cillum dolore eu fugiat nulla pariatur. Excepteur sint occaecat cupidatat non proident, sunt in culpa qui officia deserunt mollit anim id est laborum.

%% If you have bibdatabase file and want bibtex to generate the
%% bibitems, please use
%%
 \bibliographystyle{elastic-num-names} 
 \bibliography{cas-refs}

%% else use the following coding to input the bibitems directly in the
%% TeX file.

% \begin{thebibliography}{00}

% %% \bibitem{label}
% %% Text of bibliographic item

% \bibitem{}

% \end{thebibliography}
\end{document}

%% file: Introduction.tex
\section{Introduction}\label{sec:introduction}

Recently, different Graph Neural Network~(GNN) architectures have made a remarkable advance in many network learning tasks by effectively incorporating relational data in the learning process. Several GNN structures are developed to solve tasks like edge prediction, node or entire network classification, and network clustering. Rapidly, GNNs found their place in different domains where underlying relational data is available such as social network analysis, molecular biology, traffic analysis, computer vision, and recommender systems.

While in some of these domains the network structures are static and don’t change over time (such as molecular graphs) in others the network structure may change regularly. For instance, in social networks, people can build new connections or cut their previous connections, also new people could join the platform. To cope with this dynamicity, we should either train the model periodically or develop an online algorithm that could update the parameters as new changes in the network occur. Moreover, temporal patterns in network evolution could help the model better learn the dynamics and solve the tasks.

In response to this challenge, numerous methods have been proposed to utilize the time-varying nature of the network and update the model in a real-time fashion~\cite{kazemi2020representation}. Joint Dynamic User-Item Embeddings~(JODIE)~\cite{kumar2019predicting} is one of the CTDG methods that use a joint embedding update in a bipartite network setting. It updates the embeddings by iterating over the list of new edges and updates embeddings of nodes involved in each new edge. The work is mainly proposed for bipartite interaction networks that have users on one side and items on the other side, and we have the timestamped list of the interactions between users and items. To preserve the order of observing new edges in the training phase while having benefits of batching, the authors introduced t-batching. This method splits interaction time-series into batches such that we ensure each node has an up-to-date embedding, and we have seen all of its previous interactions before we process it in the current batch. By doing so, the method achieves a significant time benefit, which is helpful for training on large dynamic networks.

%In this paper, we investigate the loss function used by JODIE when utilizing t-batching for training. We show that due to having different batch sizes in t-batching, using the original Mean Squared Error~(MSE) loss will result in a bias toward interactions that end up being in smaller batches. We introduce two modified loss functions that don’t have the mentioned bias. By studying the loss function in a simplified scenario, we could analyze the outcomes theoretically, showing how the loss function’s problem would affect the accuracy of the model with a globally minimum loss function. We also examine synthetic and real-world networks to show how this bias could affect both model accuracy and the optimization process.

In this paper, we investigate the loss function used by JODIE when utilizing t-batching for training and identify a problem with the loss function. The issue is attributed to the fact that different batches created during t-batching could have varying sizes. As a consequence, using the original Mean Squared Error (MSE) loss introduces a bias, wherein batches with smaller sizes affect the loss disproportionately higher relative to their size compared to larger batches. To address this bias, we propose a solution involving reweighting of samples in each batch, based on the number of data points contained within. Consequently, we introduce two modified loss functions that eliminate this problem, where the two loss functions differ only in the weight they assign to the MSE term relative to the regularization term.

By studying the loss functions in a simplified scenario, we theoretically analyze the outcomes, illustrating how the problem with the loss function impacts the accuracy of the model with a globally minimum loss function. Furthermore, we conduct extensive examinations on various synthetic and real-world networks to demonstrate the potential effects of this bias on both model accuracy and the optimization process.

For the synthetic graph experiments, we design four groups of experiments with different network generation processes. We begin with smaller and simpler network generation processes and gradually introduce more complexity to better emulate real-world networks. These synthetic experiments serve to provide valuable insights into the impact of different loss functions on model performance.

In real-world experiments, we utilize four datasets with distinct sizes and characteristics. These datasets are derived from two-sided platforms where users interact with items such as music or posts. Additionally, we introduce a new dataset from a mobile application platform called Myket, which boasts a large number of users and items, leading to unique characteristics compared to other datasets.

%Our results show that the loss functions we proposed enable models to reach the optimal parameters more quickly. Additionally, by addressing the bias issue, we were able to attain better evaluation results compared to the original loss function.

Our extensive experiments consistently show that the proposed loss functions outperform the original one. The proposed loss functions achieve better prediction performance in almost all of the cases by addressing the bias issue. Additionally, we observe that these loss functions accelerate the optimization process, and the model finds the optima relatively faster when using the proposed loss functions.

\vspace{0.2cm}

\textbf{Contributions.}
Our contributions in this paper are threefold:
\begin{itemize}
    \item We look more in-depth at the t-batching process and indicate a problem in the original loss function. Then, we theoretically analyze the issue in a simple scenario.
    \item We introduce two loss functions that don’t suffer from the original loss function problems and show that they would be effective by running experiments on synthetic and real networks.
    \item We investigate the characteristics of a dynamic network and training process that are related to the amount of change in accuracy caused by choosing the wrong loss function.
\end{itemize}

\section{Background and Related Work}
The significance of networks lies in their ability to model scenarios involving linked entities or interactions between entities. They find wide applications in various domains, such as social networks, where users are represented as network nodes and their interactions as edges. Similarly, in molecular networks, nodes represent atoms, and edges signify the bonds between them. When dealing with networks, researchers encounter diverse practical questions.

One such question involves node or edge classification. For instance, in a social network, detecting fraudulent users based on their behavior information becomes crucial for the platform. Additionally, the question may be to classify the entire networks or sub-networks. This question arises in molecular or biological data, where the task may be to classify molecules based on their structures. Another important aspect is edge prediction, especially when modeling interactions between entities. Predicting the next interaction of each entity, known as link prediction, proves valuable in recommendation systems, aiding in the prediction of user-item interactions to provide better recommendations. Similarly, in social networks, predicting the next friendship or interaction of a user becomes relevant.

Previously, numerous approaches have been devised to tackle these questions, ranging from similarity-based heuristics to probabilistic methods. However, with the continuous advancements in machine learning techniques, novel methods have emerged to address these problems more effectively. In this paper, we delve into one such technique, specifically focusing on its loss function. The remainder of this section will explore three areas of research closely related to the setting in this paper.

\subsection{Representation Learning on Networks}
Representation learning on graphs involves transforming graph information into a vector space, where we can solve the classification or prediction tasks by using the transformed vectors in that embedding space. The book \cite{hamilton2020graph} comprehensively covers the challenges in the network domain and various techniques for representation learning on networks. These embeddings can be generated at the node, edge, or sub-graph levels, and in this work, we focus on the node-level representations.

One category of solutions for finding representations entails shallow approaches. These methods seek vector embeddings that reflect the similarity distances in the original network structure. Examples of such works include Laplacian eigenvalues, Node2Vec~\cite{grover2016node2vec}, and DeepWalk~\cite{perozzi2014deepwalk}.

Another effective approach is Graph Neural Networks (GNNs). Unlike shallow models that don't have any parameter sharing, GNNs employ different neural network structures tailored to networks, leading to improved results. A variety of neural network structures have been proposed in this area. Convolutional Graph Networks~\cite{kipf2016semi}, GraphSAGE~\cite{hamilton2017inductive}, and Graph Attention Networks\cite{velickovic2017graph} are some notable examples of GNN-based methods. A recent advancement in this field is GPT-GNN\cite{hu2020gpt}, which employs a self-supervised method for generative pre-training of the graph neural network, which significantly enhances the performance of existing methods.

The approach discussed in this paper falls under the category of graph neural network structures, utilizing Recurrent Neural Networks (RNNs) to learn node embeddings.

\subsection{Representation Learning on Dynamic Networks}
An essential aspect of network studies involves considering their variation over time. Particularly in domains like social networks, where continuous changes occur – such as the arrival of new users, events, and the formation of new connections – modeling the dynamics within the network structure becomes vital. In this area, \cite{kazemi2020representation} provides a comprehensive survey of literature and methods in this area.

The proposed methods can generally be categorized into two classes: those designed for Discrete-Time Dynamic Graphs (DTDGs) and those for Continuous-Time Dynamic Graphs (CTDGs). In the first class, some methods either disregard the timestamps of events or capture snapshots of the network structure at fixed intervals \cite{sankar2020dysat,pareja2020evolvegcn}. On the other hand, methods for CTDGs take into account event timestamps and process each event as they occur, incorporating time features of events \cite{rossi2020temporal, trivedi2019dyrep, kumar2019predicting, gao2022novel}.

In this paper, we focus on investigating the JODIE approach, a method designed for CTDGs. By handling CTDGs, JODIE is particularly well-suited for scenarios in social networks and recommendation systems where interactions occur continuously over time. The approach operates on a bipartite network setting involving users and items and dynamically updates the embeddings of both node types whenever an interaction occurs. Additionally, JODIE incorporates a module that predicts the trajectory of node movement in the embedding space, enhancing its ability to effectively incorporate temporal changes within the model.

\subsection{Determining Optimization Loss function}

%In this research, we mainly focus on the edge prediction task because that is t. While our representation learning approach can be applied to classification questions,  By addressing the edge prediction challenge, we aim to enhance the understanding and applicability of network modeling in various real-world scenarios.

The choice of the loss function plays a crucial role in machine learning methods, influencing the optimization process and the performance of trained models. Numerous studies have examined the impact of different loss functions on various tasks, highlighting their significant effects \cite{demirkaya2020exploring, li2003loss, kornblith2020s, wu2022fmd}. In the context of classification tasks, considerable attention has been given to sample weighting and its influence on model performance. Sample weighting is employed to address issues such as varying data reliabilities~\cite{hashemi2018weighted} or class imbalance. The inverse frequency weighting is commonly used to tackle imbalanced datasets \cite{wang2017learning}. Additionally, other proposals aim to control model bias towards small and large categories \cite{cui2019class, cao2019learning, shu2019meta}.

However, a study by Byrd et al. \cite{byrd2019effect} presents contrasting findings, suggesting that the choice of the loss function only affects the number of epochs required for training if the network can effectively separate the training data. The analysis also extends to cases involving regularization terms, which may lead to different behaviors. Our work aligns with this research direction on sample weighting and loss functions, as we specifically investigate the effect of sample weighting during the t-batching process and its impact on the training process and model predictions.

%The loss function is an essential element in machine learning methods that affect the optimization procedure and trained models’ performance. Previously, several studies evaluated these effects and shown how choosing the loss function could result in considerable changes in different tasks \cite{demirkaya2020exploring}\cite{li2003loss}\cite{kornblith2020s}. One direction that has got significant attention in classification tasks is how weighting different training samples could impact the model. The inverse frequency weighting is a widely used method that helps the model treat unbalanced datasets \cite{wang2017learning}. Other suggestions are also provided to control the bias of the model toward small and large categories \cite{cui2019class}\cite{cao2019learning}\cite{shu2019meta}. Opposing these experiments, Byrd et.al.\cite{byrd2019effect} indicate that if the network is capable of separating the training data, the loss function would only affect the number of epochs required for training. It also analyzes the cases where regularization terms are incorporated, which may result in different behaviors. While in this study, we don’t consider the class imbalance problem, our work is related to this line of research as we investigate the effect of sample weighting during the t-batching process and analyze its impact on the training process and model predictions.

%% file: Setup.tex
\section{Problem Setup}
% \begin{itemize}
%     \item JODIE description
%     \item tbatching & loss functions
%     % \item theory
% \end{itemize}
%This section first goes through the JODIE method and then investigates the t-batching process in detail. After going through the technique, we observe the original loss function used in the t-batching process and introduce our two alternative loss functions.

Within this section, we will outline the mathematical notation for the problem and introduce the JODIE approach. We will further investigate the t-batching approach used by JODIE in order to enhance the learning process. Finally, we will highlight a discrepancy in the loss function of t-batching concerning the desired loss, prompting the introduction of two fixed loss functions as alternatives.

Consider a bipartite network setting comprising a user set $\mathcal{U}$ and an item set $\mathcal{I}$. Additionally, there exists a set $\mathcal{S}$ that contains information about interactions between users and items that we use for the training. Each interaction in $\mathcal{S}$ is represented as a tuple with four elements: $(u, j, f, t)$. Here, $u \in \mathcal{U}$ corresponds to the user, $j \in \mathcal{I}$ to the item, $f \in \mathbb{R}^w$ to a set of features associated with the interaction, and $t$ denotes the time at which the event occurs.

The objective is to discover representations of nodes residing in $\mathbb{R}^d$ at each time, capturing the evolving information about the node's status in the network. We denote $u(t) \in \mathbb{R}^d$ as the embedding of user $u$ at time $t$, and similarly, $i(t)$ represents the embedding of item $i$ at time $t$. These embeddings serve as informative snapshots that reflect the current state of the users and items in the network.

%JODIE model overview is illustrated in Figure \ref{fig:jodie-overview}. It uses a joint representation update mechanism to alter embeddings of users and items involved in new edges. Two joint RNN structures update users and items embeddings, respectively, using the edge’s features. JODIE also introduces a projection layer that could project user embeddings, enabling the model to predict the user embedding for future times. Finally, the model uses both the projected user embedding and embedding of the last item user has interacted with to predict the next item’s embedding. With having the next item’s prediction, we could update model parameters by having a loss function comparing the predictions to ground-truth interactions observed in the data. The goal of JODIE is to optimize

Let's delve into JODIE, a model that utilizes interaction information to construct and update node embeddings. Figure \ref{fig:jodie-overview} provides an overview of the model. JODIE employs a joint representation update mechanism, modifying embeddings of users and items involved in new edges. Two joint RNN structures update user and item embeddings, respectively, using the edge's features. Additionally, JODIE introduces a projection layer, enabling the model to predict future user embeddings.

To predict the embedding of the next item, the model utilizes both the projected user embedding and the embedding of the last item the user interacted with. The model's parameters are updated through a loss function that compares these predictions with ground-truth interactions observed in the data. The loss function for the JODIE method is defined as:

\begin{equation}
\begin{aligned}
loss_{JODIE} = \sum_{(u, j, t, f) \in \mathcal{S}}& 
 ||\hat{j}(t) - j(t^-)||_2^2\\
& + \lambda_U ||u(t) - u(t^-)||_2^2 \\
& + \lambda_I||j(t) - j(t^-)||_2^2  
\label{eq:jodie-loss}
\end{aligned}
\end{equation}

Here, $u(t)$ and $j(t)$ represent the updated embeddings of the user and item involved in the interaction, while $u(t^-)$ and $j(t^-)$ correspond to their embeddings right before the interaction. The term $\hat{j(t)}$ refers to the predicted embedding of the item that the user $u$ is expected to interact with.

The first term in the loss function is the prediction loss, aiming to minimize the distance between the predicted item embedding and the actual item with which the user has interacted, as observed in the data. The last two terms serve as regularization terms, preventing the model from updating the embeddings after each interaction by a large magnitude. By doing so, the embedding updates become smoother, retaining long-term information more effectively, and preserving past interactions' valuable information.

%where $u(t)$ and $j(t)$ are the updated embeddings of the user and item involved in the interaction, and $u(t^-)$ and $j(t^-)$ are their embeddings before the interaction.The $\hat{j(t)}$ is the embedding of the item that user $u$ is going to interact with. The first two terms in this loss function perform as regularization terms that prevent the model from updating the embeddings after each interaction by a large magnitude. In this way, the embedding updates are more smooth and could hold long-term information better, by keeping information from the past. The final term is the prediction loss which tries to minimize the distance between the predicted item embedding and the item that the user has interacted with from the data.

\input{figure}

To ensure that the embeddings are up-to-date and contain all relevant historical information, it is necessary to process the interactions in the same order they occur in the data. One approach to achieving this is to process the training set one interaction at a time. However, this can be time-consuming due to the lack of batching. On the other hand, using a simple batching approach on the interaction sequence may lead to batches with repetitive nodes. This presents a problem, as having a node appear multiple times in a batch would mean using the same version of its embedding that was last updated before encountering that batch. Updating embeddings for repetitive nodes can also be challenging, as there would be multiple choices for the new embeddings.

To overcome these challenges, Kumar et al.~\cite{kumar2019predicting} proposed a method that efficiently benefits from batching while addressing the issues mentioned above. We will next delve deeper into this method and its implementation.

%As we need the up-to-date embedding of nodes for each of the predictions, we should ensure that each node’s interactions are processed with the original order observed in the data. One idea is to process the training set one interaction at a time. While this approach satisfies the model requirement, it would be prolonged as we are not using any type of batching. Moreover, using a simple batching on interaction sequence may result in batches that have repetitive nodes. Having a node twice or more in a batch would be problematic as for all of its interactions, we are using the same version of the embedding that is last updated before seeing this batch. It may also cause challenges in updating the repetitive node’s embedding as we would have multiple choices for its new embedding. T-batching proposed by Kumar et al.\cite{kumar2019predicting} provides a method that could benefit from batching while overcoming these challenges.

\subsection{T-batching}

The t-batching process begins by initializing a set of empty batches. Subsequently, we enumerate over all interactions and add each interaction to the first batch that shares no common nodes with the interaction. The detailed process is described in Algorithm \ref{alg:tbatching}. The variable $U_u$ indicates the number of the last batch in which user $u$ is placed, and similarly, $I_j$ represents the number of the last batch in which item $j$ is placed.

%The process starts with initializing a set of empty batches. We then append each interaction to the first batch with no common nodes. The process is stated in Algorithm \ref{alg:tbatching}. $U_i$ and $I_i$ indicate the number of the last batch this user or item is placed in.

\input{algorithm}

%Figure \ref{fig:tbatch-example} shows a simple example of how t-batching works. In this network, edges arrive in order as indicated in their labels. The process starts with empty batches. By seeing the first edge, the method appends it to batch \#1. The second edge could also go to this batch as it does not share any nodes with edges in the batch. For the third edge, we have to append it to a new batch as it has common nodes with both edges in batch \#1.

\begin{figure}[!t]
    \centering
    \includegraphics[width=0.8\textwidth]{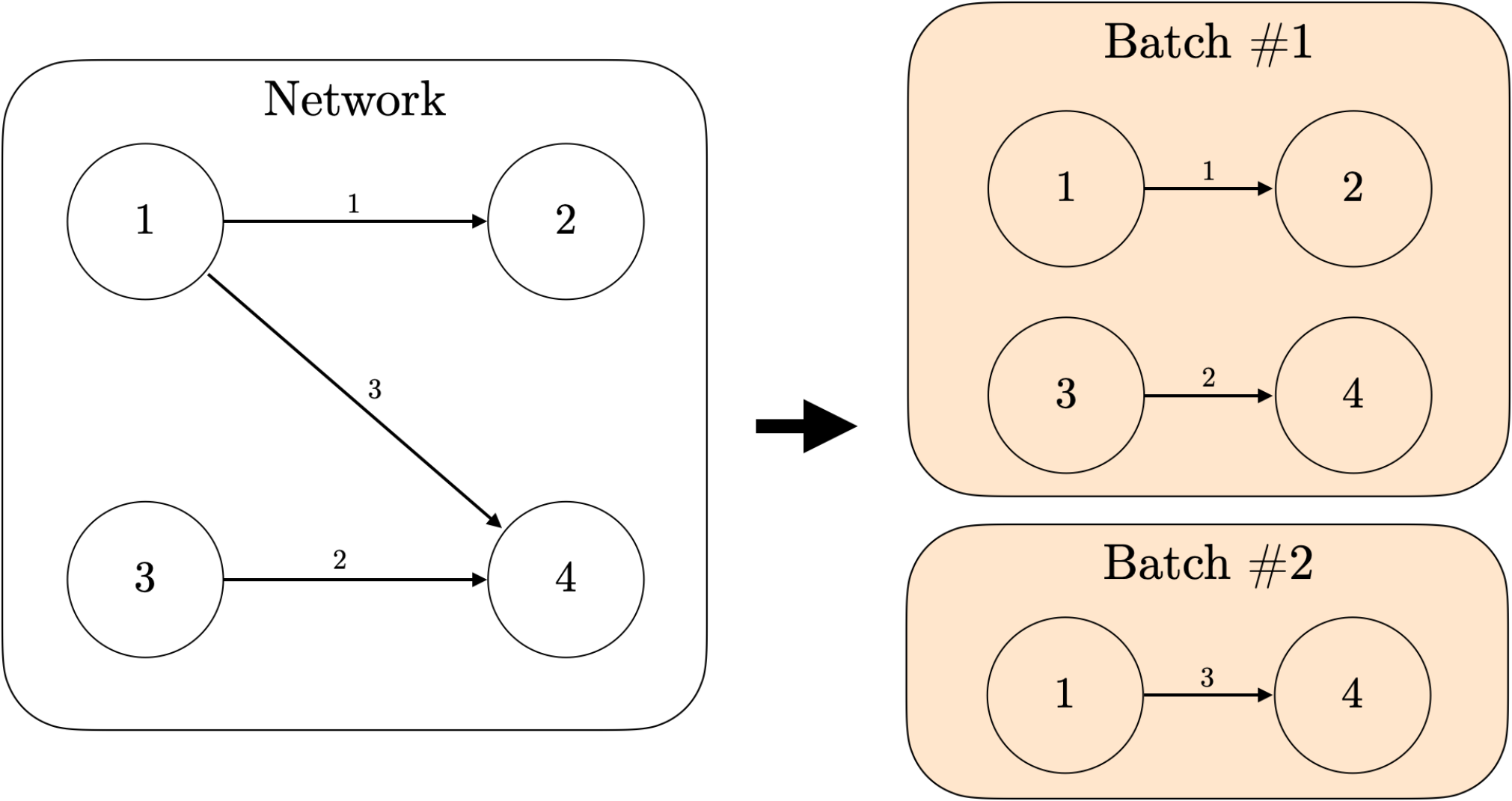}
    \caption{An example of t-batching process}
    \label{fig:tbatch-example}
\end{figure}

To provide a visual representation of the t-batching method, Figure \ref{fig:tbatch-example} illustrates a simple example involving a four-node network. The process begins with empty batches, and edges arrive in the order indicated by their labels. Upon encountering the first edge, the method appends it to batch \#1. The second edge can also be added to this batch, as it does not share any nodes with the existing edges in the batch. However, for the third edge, a new batch is created, as it shares common nodes with both edges in batch \#1.

%This example shows that t-batching may result in different batch sizes, which we state could cause problems when we use inappropriate loss function. 
%JODIE uses

Upon incorporating t-batching, JODIE employs the following loss function:

\begin{equation}
\begin{aligned}
loss_{tbatch} = \sum_{\mathcal{S}_b \in \mathcal{B}} \biggl( &\frac{1}{|\mathcal{S}_b|d} \sum_{(u, j, t, f) \in \mathcal{S}_b} ||\hat{j}(t) - j(t^-)||_2^2\\
&+ \frac{1}{|\mathcal{U}|d}\sum_{u \in \mathcal{U}}\lambda_U ||u - u^-||_2^2 \\
&+ \frac{1}{|\mathcal{I}|d}\sum_{j \in \mathcal{I}}\lambda_I||j - j^-||_2^2  \biggr)
\label{eq:jodie-loss-batch}
\end{aligned}
\end{equation}

where $\mathcal{B}$ is the set of batches that is a partitioning of all interactions in $\mathcal{S}$, and each batch $\mathcal{S}_b \in \mathcal{B}$ is a subset of $\mathcal{S}$. The first term corresponds to the MSE loss, comparing the predictions in the batch to the corresponding ground-truth observations. The last two terms serve as regularization losses, comparing embeddings of all nodes before and after processing the t-batch.

%as its loss function when using t-batching and calculates each batch’s loss separately. Here, $d$ is the dimension of user and item embeddings. The first two terms are MSE losses comparing embeddings of all nodes before and after processing this t-batch. The final term is also an MSE loss, comparing all of the predictions to all of the ground-truth observations. This equation differs from the equation \ref{eq:jodie-loss} in two aspects. First, by using MSE on the regularization terms, we have a rescaling in the regularization factor. Second and more importantly, we have $|\mathcal{S}_b|$ in the third term’s multiplier, which reduces the contribution of interactions in larger batches.

\subsection{Problem with the Loss Function and Proposed Solutions}

As observed in the example in Figure \ref{fig:tbatch-example}, t-batching may lead to varying batch sizes, which can be problematic when using $loss_{tbatch}$. Let's compare the $loss_{tbatch}$ function to the original JODIE loss function, $loss_{JODIE}$. Equation \ref{eq:jodie-loss-batch} differs from Equation \ref{eq:jodie-loss} in two aspects. First, by employing the MSE on the regularization terms, we introduce a rescaling in the regularization factor. Second, and more significantly, the inclusion of $|\mathcal{S}_b|$ in the multiplier of the third term results in a reduced contribution of interactions in larger batches.

To address this issue, we propose two modified loss functions that remove $|\mathcal{S}_b|$ from the first term multiplier. These are:

% \begin{align*}
% loss_{item-sum} &= \sum_{\mathcal{S}_b \in batches} \frac{1}{|\mathcal{U}|d}\sum_{u \in \mathcal{U}}\lambda_U ||u - u^-||_2^2 \\
% &+ \frac{1}{|\mathcal{I}|d}\sum_{j \in \mathcal{I}}\lambda_I||j - j^-||_2^2 \\
% &+ \frac{1}{d} \sum_{(u, j, t, f) \in \mathcal{S}_b} ||\hat{j}(t) - j(t^-)||_2^2
% \end{align*}

\begin{equation}
\begin{aligned}
loss_{item-sum} = \sum_{\mathcal{S}_b \in \mathcal{B}} \biggl( & \frac{1}{d} \sum_{(u, j, t, f) \in \mathcal{S}_b} ||\hat{j}(t) - j(t^-)||_2^2  \\
&+ \frac{1}{|\mathcal{U}|d}\sum_{u \in \mathcal{U}}\lambda_U ||u - u^-||_2^2\\
&+ \frac{1}{|\mathcal{I}|d}\sum_{j \in \mathcal{I}}\lambda_I||j - j^-||_2^2 
\biggr)
\end{aligned}
\label{eq:jodie-loss-batch-item-sum}
\end{equation}

and

\begin{equation}
\begin{aligned}
loss_{full-sum} = \sum_{\mathcal{S}_b \in \mathcal{B}}& \biggl( 
\sum_{(u, j, t, f) \in \mathcal{S}_b} ||\hat{j}(t) - j(t^-)||_2^2 \\
&+ \frac{1}{|\mathcal{U}|d}\sum_{u \in \mathcal{U}}\lambda_U ||u - u^-||_2^2\\
&+ \frac{1}{|\mathcal{I}|d}\sum_{j \in \mathcal{I}}\lambda_I||j - j^-||_2^2 \biggr).  
\end{aligned}
\label{eq:jodie-loss-batch-full-sum}
\end{equation}

The functions differ only in the $\frac{1}{d}$ multiplier, which, if we have the same regularization factors, results in the importance of regularization being lower in $loss_{full-sum}$ compared to $loss_{item-sum}$. By adopting these modified loss functions, we can mitigate the bias introduced by t-batching and achieve more accurate and stable updates in the node embeddings.

%To overcome the mentioned problem, we introduce two modified loss functions, which remove the $|\mathcal{S}_b|$ from the third term multiplier. These two are

% and

% These two differ only on the $\frac{1}{d}$ multiplier, which in case we have the same regularization factors makes the importance of regularization lower in the second function.

%% file: figure.tex
\begin{figure}[!t]
    \centering
    
\resizebox{0.9\textwidth}{!}{    
\begin{tikzpicture}[x=0.75pt,y=0.75pt,yscale=-1,xscale=1]
%uncomment if require: \path (0,362); %set diagram left start at 0, and has height of 362

%Flowchart: Alternative Process [id:dp8328435712832335] 
\draw  [fill={rgb, 255:red, 6; green, 126; blue, 196 }  ,fill opacity=1 ] (23,92.6) .. controls (23,89.51) and (25.51,87) .. (28.6,87) -- (50.9,87) .. controls (53.99,87) and (56.5,89.51) .. (56.5,92.6) -- (56.5,113.4) .. controls (56.5,116.49) and (53.99,119) .. (50.9,119) -- (28.6,119) .. controls (25.51,119) and (23,116.49) .. (23,113.4) -- cycle ;
%Flowchart: Connector [id:dp6083022035707272] 
\draw  [fill={rgb, 255:red, 2; green, 111; blue, 1 }  ,fill opacity=1 ] (24,222.5) .. controls (24,213.94) and (30.94,207) .. (39.5,207) .. controls (48.06,207) and (55,213.94) .. (55,222.5) .. controls (55,231.06) and (48.06,238) .. (39.5,238) .. controls (30.94,238) and (24,231.06) .. (24,222.5) -- cycle ;
%Shape: Rectangle [id:dp21133672851156537] 
\draw  [fill={rgb, 255:red, 105; green, 2; blue, 2 }  ,fill opacity=1 ] (59.5,141) -- (66.5,141) -- (66.5,190) -- (59.5,190) -- cycle ;
%Shape: Rectangle [id:dp7852361469781193] 
\draw  [fill={rgb, 255:red, 6; green, 126; blue, 196 }  ,fill opacity=1 ] (151.5,62) -- (201.5,62) -- (201.5,152) -- (151.5,152) -- cycle ;
%Shape: Rectangle [id:dp08582012291946584] 
\draw  [fill={rgb, 255:red, 2; green, 111; blue, 1 }  ,fill opacity=1 ] (152.5,196) -- (202.5,196) -- (202.5,286) -- (152.5,286) -- cycle ;
%Curve Lines [id:da4370648673479137] 
\draw [color={rgb, 255:red, 6; green, 126; blue, 196 }  ,draw opacity=1 ]   (57.5,104) .. controls (100.07,94.1) and (88.73,94.98) .. (149.63,92.09) ;
\draw [shift={(151.5,92)}, rotate = 537.27] [color={rgb, 255:red, 6; green, 126; blue, 196 }  ,draw opacity=1 ][line width=0.75]    (10.93,-3.29) .. controls (6.95,-1.4) and (3.31,-0.3) .. (0,0) .. controls (3.31,0.3) and (6.95,1.4) .. (10.93,3.29)   ;

%Curve Lines [id:da9833296035353448] 
\draw [color={rgb, 255:red, 6; green, 126; blue, 196 }  ,draw opacity=1 ]   (57.5,104) .. controls (100.07,94.1) and (86.77,220.43) .. (147.63,220.05) ;
\draw [shift={(149.5,220)}, rotate = 537.27] [color={rgb, 255:red, 6; green, 126; blue, 196 }  ,draw opacity=1 ][line width=0.75]    (10.93,-3.29) .. controls (6.95,-1.4) and (3.31,-0.3) .. (0,0) .. controls (3.31,0.3) and (6.95,1.4) .. (10.93,3.29)   ;

%Curve Lines [id:da08980567878479717] 
\draw [color={rgb, 255:red, 105; green, 2; blue, 2 }  ,draw opacity=1 ]   (68.5,167) .. controls (111.07,157.1) and (86.99,119.76) .. (147.64,116.1) ;
\draw [shift={(149.5,116)}, rotate = 537.27] [color={rgb, 255:red, 105; green, 2; blue, 2 }  ,draw opacity=1 ][line width=0.75]    (10.93,-3.29) .. controls (6.95,-1.4) and (3.31,-0.3) .. (0,0) .. controls (3.31,0.3) and (6.95,1.4) .. (10.93,3.29)   ;

%Curve Lines [id:da2056995047768262] 
\draw [color={rgb, 255:red, 105; green, 2; blue, 2 }  ,draw opacity=1 ]   (68.5,167) .. controls (111.07,157.1) and (87.97,242.27) .. (148.64,241.06) ;
\draw [shift={(150.5,241)}, rotate = 537.27] [color={rgb, 255:red, 105; green, 2; blue, 2 }  ,draw opacity=1 ][line width=0.75]    (10.93,-3.29) .. controls (6.95,-1.4) and (3.31,-0.3) .. (0,0) .. controls (3.31,0.3) and (6.95,1.4) .. (10.93,3.29)   ;

%Curve Lines [id:da7034506278396155] 
\draw [color={rgb, 255:red, 2; green, 111; blue, 1 }  ,draw opacity=1 ]   (55,222.5) .. controls (97.57,212.6) and (86.72,263.96) .. (147.63,262.07) ;
\draw [shift={(149.5,262)}, rotate = 537.27] [color={rgb, 255:red, 2; green, 111; blue, 1 }  ,draw opacity=1 ][line width=0.75]    (10.93,-3.29) .. controls (6.95,-1.4) and (3.31,-0.3) .. (0,0) .. controls (3.31,0.3) and (6.95,1.4) .. (10.93,3.29)   ;

%Curve Lines [id:da8305278671426575] 
\draw [color={rgb, 255:red, 2; green, 111; blue, 1 }  ,draw opacity=1 ]   (55,222.5) .. controls (97.57,212.6) and (87.7,138.5) .. (148.63,134.11) ;
\draw [shift={(150.5,134)}, rotate = 537.27] [color={rgb, 255:red, 2; green, 111; blue, 1 }  ,draw opacity=1 ][line width=0.75]    (10.93,-3.29) .. controls (6.95,-1.4) and (3.31,-0.3) .. (0,0) .. controls (3.31,0.3) and (6.95,1.4) .. (10.93,3.29)   ;

%Straight Lines [id:da9598149015531282] 
\draw    (39.5,120) -- (39.5,206) ;

%Curve Lines [id:da4634302138447983] 
\draw [color={rgb, 255:red, 6; green, 126; blue, 196 }  ,draw opacity=1 ]   (204.5,105) .. controls (247.5,95) and (267.5,38) .. (206.5,33) .. controls (146.11,28.05) and (42.6,38.78) .. (40.53,85.57) ;
\draw [shift={(40.5,87)}, rotate = 270] [color={rgb, 255:red, 6; green, 126; blue, 196 }  ,draw opacity=1 ][line width=0.75]    (10.93,-3.29) .. controls (6.95,-1.4) and (3.31,-0.3) .. (0,0) .. controls (3.31,0.3) and (6.95,1.4) .. (10.93,3.29)   ;

%Curve Lines [id:da4211615995304979] 
\draw [color={rgb, 255:red, 2; green, 111; blue, 1 }  ,draw opacity=1 ]   (205.5,243) .. controls (248.5,233) and (259.5,325) .. (185.5,330) .. controls (112.61,334.93) and (45.54,279.7) .. (39.72,239.81) ;
\draw [shift={(39.5,238)}, rotate = 444.29] [color={rgb, 255:red, 2; green, 111; blue, 1 }  ,draw opacity=1 ][line width=0.75]    (10.93,-3.29) .. controls (6.95,-1.4) and (3.31,-0.3) .. (0,0) .. controls (3.31,0.3) and (6.95,1.4) .. (10.93,3.29)   ;

%Shape: Rectangle [id:dp7381018522867135] 
\draw   (285.5,197) -- (369.5,197) -- (369.5,278) -- (285.5,278) -- cycle ;
%Curve Lines [id:da4051734866479647] 
\draw [color={rgb, 255:red, 2; green, 111; blue, 1 }  ,draw opacity=1 ]   (205.5,243) .. controls (248.07,233.1) and (221.05,238.88) .. (281.64,236.09) ;
\draw [shift={(283.5,236)}, rotate = 537.27] [color={rgb, 255:red, 2; green, 111; blue, 1 }  ,draw opacity=1 ][line width=0.75]    (10.93,-3.29) .. controls (6.95,-1.4) and (3.31,-0.3) .. (0,0) .. controls (3.31,0.3) and (6.95,1.4) .. (10.93,3.29)   ;

%Straight Lines [id:da62421078134318] 
\draw    (331.5,315) -- (331.5,280) ;
\draw [shift={(331.5,278)}, rotate = 450] [color={rgb, 255:red, 0; green, 0; blue, 0 }  ][line width=0.75]    (10.93,-3.29) .. controls (6.95,-1.4) and (3.31,-0.3) .. (0,0) .. controls (3.31,0.3) and (6.95,1.4) .. (10.93,3.29)   ;

%Straight Lines [id:da6124353412502923] 
\draw    (369.5,235) -- (449.5,235) ;
\draw [shift={(451.5,235)}, rotate = 180] [color={rgb, 255:red, 0; green, 0; blue, 0 }  ][line width=0.75]    (10.93,-3.29) .. controls (6.95,-1.4) and (3.31,-0.3) .. (0,0) .. controls (3.31,0.3) and (6.95,1.4) .. (10.93,3.29)   ;

%Shape: Rectangle [id:dp659440069676605] 
\draw   (455,106) -- (549.5,106) -- (549.5,262) -- (455,262) -- cycle ;
%Curve Lines [id:da000378862159404858] 
\draw [color={rgb, 255:red, 6; green, 126; blue, 196 }  ,draw opacity=1 ]   (204.5,105) .. controls (247.07,95.1) and (386.67,177.33) .. (450.59,176.06) ;
\draw [shift={(452.5,176)}, rotate = 537.27] [color={rgb, 255:red, 6; green, 126; blue, 196 }  ,draw opacity=1 ][line width=0.75]    (10.93,-3.29) .. controls (6.95,-1.4) and (3.31,-0.3) .. (0,0) .. controls (3.31,0.3) and (6.95,1.4) .. (10.93,3.29)   ;

%Straight Lines [id:da8799105789630551] 
\draw    (550,192) -- (623.5,192) ;
\draw [shift={(625.5,192)}, rotate = 180] [color={rgb, 255:red, 0; green, 0; blue, 0 }  ][line width=0.75]    (10.93,-3.29) .. controls (6.95,-1.4) and (3.31,-0.3) .. (0,0) .. controls (3.31,0.3) and (6.95,1.4) .. (10.93,3.29)   ;

% Text Node
\draw (40.5,102) node    {$i$};
% Text Node
\draw (40.5,221.5) node    {$u$};
% Text Node
\draw (46,163) node    {$f$};
% Text Node
\draw (176.5,107) node    {$RNN_{i}$};
% Text Node
\draw (177.5,239) node    {$RNN_{u}$};
% Text Node
\draw (121,279) node    {$u\left( t^{-}\right)$};
% Text Node
\draw (140,159) node    {$u\left( t^{-}\right)$};
% Text Node
\draw (124,77) node    {$i\left( t^{-}\right)$};
% Text Node
\draw (139,196) node    {$i\left( t^{-}\right)$};
% Text Node
\draw (254,278) node    {$u( t)$};
% Text Node
\draw (261,84) node    {$i( t)$};
% Text Node
\draw (327.5,237.5) node   [align=left] {Projection\\ \ \ \ Layer};
% Text Node
\draw (330,326) node    {$\Delta $};
% Text Node
\draw (410,252) node    {$\hat{u}( t\ +\ \Delta )$};
% Text Node
\draw (502.25,184) node   [align=left] {Prediction\\ \ \ \ Layer};
% Text Node
\draw (590,213) node    {$\hat{i}( t\ +\ \Delta )$};

\end{tikzpicture}}
    \caption{JODIE model overview}
    \label{fig:jodie-overview}
\end{figure}

%% file: algorithm.tex
\begin{algorithm}
\caption{T-batching procedure}
\label{alg:tbatching}
\begin{algorithmic}
\REQUIRE Array $\mathcal{S}$ consisting $(u, j, f, t)$ tuples, ordered with time
	\STATE $U_u \leftarrow -1 \quad \forall u \in \mathcal{U}$
	\STATE $I_j \leftarrow -1 \quad \forall j \in \mathcal{I}$
	\STATE Start with empty batches
	\FOR{$(u, j, f, t) \in \mathcal{S}$}
		\STATE $k \leftarrow max(U_u + 1, I_j + 1)$
		\STATE Add interaction $e$ to batch $k$
		\STATE $U_u, I_j \leftarrow k$
	\ENDFOR
	\STATE Run optimization on the batches sequentially
\end{algorithmic}
\end{algorithm}

%% file: Theory.tex
\section{Theoretical Analysis in a Simplified Scenario}

In this section, we present a theoretical analysis of how the original loss function may introduce a bias that impacts the model's accuracy. As observed in the previous section, the t-batching process can lead to batches with varying sizes. To investigate this further, we propose a random process for generating dynamic networks, where edges of the same type are always present in batches with specific sizes during t-batching. This approach allows us to analyze the effect of batch sizes on predictions and model performance. The detailed network generation process is described in Algorithm \ref{alg:graph-generation}.

%In this section, we theoretically demonstrate how the original loss function may result in a bias that could affect the model’s accuracy. As seen in the previous section, the t-batching process may end up having batches with different sizes. We introduce a random process for generating dynamic networks that, during t-batching, edges of the same type would always be in batches with specific sizes. This way, we could analyze the effect of batch sizes in predictions. The network generation process is described in algorithm \ref{alg:graph-generation}.

\input{generation-algo}

This method generates dynamic networks involving two users, two items, and $k \in 2\mathbb{N}$ timestamped interactions between users and items. The generation process comprises $\frac{k}{2}$ rounds. In each round, we establish one deterministic edge between user 1 and item 2, and a Bernoulli trial with parameter $p$ selects the second edge for user 3 from items 2 and 4.

\begin{figure}[!t]
    \centering
    \includegraphics[width=0.7\textwidth]{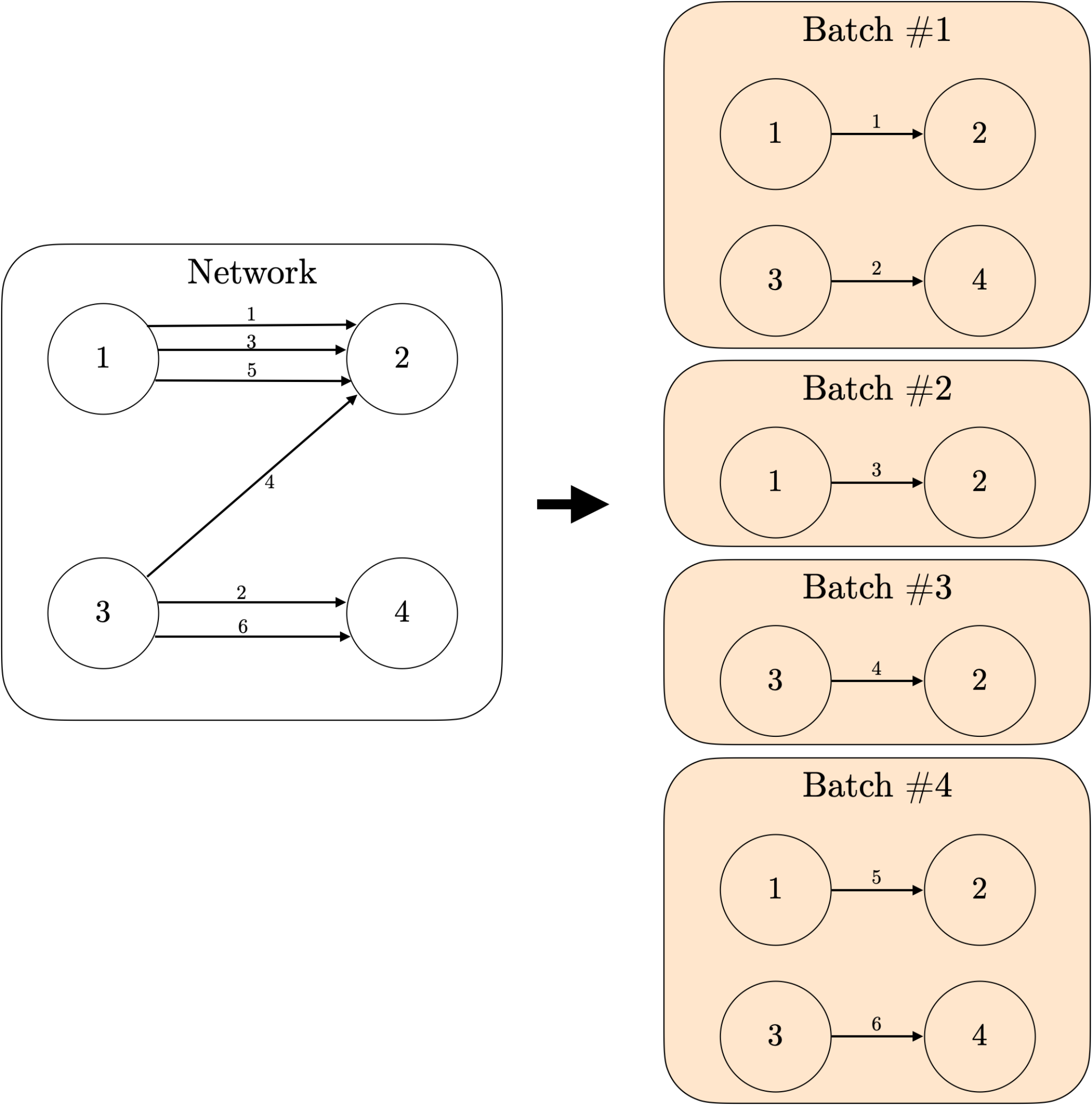}
    \caption{An example network generated by Algorithm \ref{alg:graph-generation}, and how the t-batching would create batches.}
    \label{fig:generated-example}
\end{figure}

During t-batching, the process initiates by placing the (1, 2) edge in an empty batch. Subsequently, if the next edge corresponds to (3, 4), it can be added to the same batch; however, if it corresponds to (3, 2), it must be assigned to a new batch. The subsequent (1, 2) edge should also go into a new batch, as in both cases, it would share nodes with the previous batch. If the previous batch contains (1, 2) and (3, 4), it shares nodes with (1, 2), also if it only contains (3, 2), having item 2 in common prevents them from being grouped together. As a result, (3, 2) edges always fit into a batch with one item, whereas (3, 4) edges end up in a batch with two items. An illustrative example of this process is presented in Figure \ref{fig:generated-example}.

% This method creates dynamic networks with two users, two items, and $k \in 2\mathbb{N}$ timestamped interactions between users and items. The generation consists of $\frac{k}{2}$ rounds. In each round, we will have one deterministic edge between user 1 and item 2, and a Bernoulli trial with parameter p selects the second edge for user 3 from items 2 and 4. During t-batching, we will always start with putting the (1, 2) edge in an empty batch. Then, if the next edge would be (3, 4) we could put it in the same batch, and if it would be (3, 2) we have to put it in a new batch. The next (1, 2) edge should be again in a new batch because, in each of the cases, it would have common nodes with the previous batch. If the previous batch consists of (1, 2), and (3, 4) it will have common nodes with (1, 2), and if it consists only of (3, 2), having item 2 in common prevents them from joining. This way, (3, 2) edges will always fit in a batch with one item, and (3, 4) edges end up being in a batch with two items. An example of this process is illustrated in figure \ref{fig:generated-example}.

% First, we try the optimal edge prediction model for network samples generated by this process. The prediction for user 1 is fairly easy as it will always interact with item 2. However, the case for user 3 may be challenging. If $k$ would be large enough, by the first law of large number we could assume the number of (3, 4) edges would be $\approx p\frac{k}{2}$ and the number of (3, 2) edges would be $\approx (1 - p)\frac{k}{2}$. Also, as the generation process does not use any memory and is time-independent, the model can not achieve any information from sequence patterns that could enhance its prediction for user 3. Knowing the parameter $p$, the optimal model should always predict the same item for user 3 that would result in more accurate predictions. So the optimal model predictions should be

First, we evaluate the optimal edge prediction model on network samples generated by this process. Predicting the interaction for user 1 is relatively straightforward since it will always interact with item 2. However, the case for user 3 may pose a challenge. If $k$ is sufficiently large, the law of large numbers allows us to approximate the number of (3, 4) edges to be around $p\frac{k}{2}$ and the number of (3, 2) edges to be approximately $(1 - p)\frac{k}{2}$. Additionally, as the generation process lacks memory and is time-independent, the model cannot gain any information from sequence patterns to enhance its prediction for user 3. Given the parameter $p$, the optimal model should consistently predict the same item for user 3, leading to more accurate predictions. Thus, the optimal model's predictions can be expressed as follows:

\begin{equation}
\begin{aligned}
prediction(1) &= 2\\
prediction(3) &= \begin{cases}
2 & p \leq 0.5\\
4 & p > 0.5
\end{cases}.
\end{aligned}
\end{equation}

We now proceed to examine the learning process utilizing the loss function $loss_{tbatch}$ for our analysis. For simplicity, we temporarily exclude the regularization terms from our investigation in this section, resulting in a simplified loss function, as expressed in Equation \ref{eq:jodie-loss-batch-simple}:

%Now we investigate the learning process with $loss_{tbatch}$ as loss function. For simplicity, we don’t consider the regularization terms in the analyzes of this section. So the loss function would be like

\begin{equation}
\begin{aligned}
loss_{tbatch-simple} &=& \sum_{\mathcal{S}_b \in \mathcal{B}}\frac{1}{|\mathcal{S}_b|d} \sum_{(u, j, t, f) \in \mathcal{S}_b} ||\hat{j}(t) - j(t^-)||_2^2
\end{aligned}
\label{eq:jodie-loss-batch-simple}
\end{equation}

%As the prediction for user 1 is trivial, we only consider edges for user 3 in the loss function 
Given that predicting interactions for user 1 is straightforward, we solely focus on edges concerning user 3 in the loss function. This would result in the refined representation of the loss function as:

\begin{equation}
\begin{aligned}
loss_{tbatch-simple} =  \sum_{\mathcal{S}_b \in \mathcal{B}}\biggl(&\sum_{(3, 4, t, f)}\frac{1}{2d} ||\hat{j_3}(t) - j_4(t^-)||_2^2 \\
& + \sum_{(3, 2, t, f)}\frac{1}{d} ||\hat{j_3}(t) - j_2(t^-)||_2^2\biggr)
\end{aligned}
\label{eq:jodie-loss-batch-simple-rewrite}
\end{equation}

%The multiplier before (3, 4) edges have an extra $\frac{1}{2}$ term because they will be in batches of size two. Now if we assume we have $\frac{k}{2}$ edges for user 3 in the training and k is large enough to use the law of large number we could rewrite $loss_{simple}$ as

The multiplier before (3, 4) edges has an additional $\frac{1}{2}$ term since they are placed in batches of size two. Assuming we have $\frac{k}{2}$ edges for user 3 during training, and considering k is sufficiently large to employ the first law of large numbers, we can rewrite $loss_{tbatch-simple}$ as:

\begin{equation}
\begin{aligned}
loss_{tbatch-simple} \approx  & \frac{kp}{2} \frac{1}{2d} ||\hat{j_3}(t) - j_4(t^-)||_2^2\\
& + \frac{k(1 - p)}{2} \frac{1}{d} ||\hat{j_3}(t) - j_2(t^-)||_2^2
\end{aligned}
\label{eq:jodie-loss-batch-simple-rewrite2}
\end{equation}

%To minimize this loss function, assuming the embeddings of items are static in time, the model should choose the position of the predicted embedding to be either equal to item 2 or 4. To do so, the model should consider which multiplier would be higher so that minimizing that term would have a greater impact on the loss. So it’s decision boundary would be calculated as

To minimize this loss function, assuming the embeddings of items are static in time, the model should choose the position of the predicted embedding to be either equal to item 2 or 4. It needs to consider which multiplier would yield a higher value so that minimizing that term would have a greater impact on the loss. Consequently, the decision boundary can be calculated as:

\begin{equation}
\begin{aligned}
\frac{kp}{2}\frac{1}{2} &< \frac{k(1 - p)}{2}\\
\implies p &< 2(1 - p)\\
\implies p &< \frac{2}{3}
\end{aligned}
\label{eq:jodie-loss-batch-simple-criterion}
\end{equation}

We observe that the boundary differs from the optimal model's boundary, which is $\frac{1}{2}$. Thus, when $\frac{1}{2} < p < \frac{2}{3}$, the model with loss function $loss_{tbatch-simple}$ would predict item 4, which is accurate for $1 - p$ cases. However, due to the constraint $\frac{1}{2} < p$, it will have an accuracy lower than the optimal accuracy $p$, which the model could achieve by predicting item 3. By similar calculations, it can be demonstrated that other provided loss functions do not exhibit this issue.

%We see that the boundary differs from the boundary of optimal model which is $\frac{1}{2}$. This means when we have $\frac{1}{2} < p < \frac{2}{3}$ the model with loss function $loss_{simple}$ would predict item 4 that would be a correct prediction for $1 - p$ cases. Because we have $\frac{1}{2} < p$ it will be lower than the optimum accuracy $p$ the model can achieve with predicting item 3. With the same calculations, it is easy to show that other provided loss functions don’t suffer from this problem.

%% file: generation-algo.tex
\begin{algorithm}
\caption{Simple graph generation method for theoretical analysis}
\label{alg:graph-generation}
\begin{algorithmic}
\REQUIRE $k \in 2\mathbb{N}$: number of edges, $p$ Bernoulli trial parameter
	\STATE Start with empty set of edges
	\FOR{$i=0 \rightarrow \frac{k}{2}$}
		\STATE Add edge $(1, 2)$
		\STATE $r \sim Uniform(0, 1)$
		\IF{$r < p$}
			\STATE Add edge $(3, 4)$
		\ELSE
			\STATE Add edge $(3, 2)$
		\ENDIF
	\ENDFOR
\end{algorithmic}
\end{algorithm}

%% file: Results.tex
\section{Experiments \& Results}
% \begin{itemize}
%     \item Synthetic
%     \item Real world
% \end{itemize}
%With different experiments on synthetic and real-world dynamic networks, we show how this problem in the loss function would affect the training process and accuracy. Experiments are divided into four groups, three experiments on synthetic networks and one on real networks.

Through a series of diverse experiments conducted on both synthetic and real-world dynamic networks, we present a comprehensive analysis of how the aforementioned issue in the loss function impacts the training process and accuracy. Furthermore, we demonstrate that the proposed alternative loss functions effectively address this problem. Our experiments are categorized into five distinct groups, consisting of four experiments on synthetic networks and one on real networks.
We commence by exploring straightforward network generation processes and gradually advancing towards more intricate structures to better emulate real-world networks more. In the real-world experiment, we evaluate the methods using four distinct interaction networks, each exhibiting unique characteristics. Notably, we introduce a novel dataset encompassing installation interactions in an Android application market platform (MyKet), which constitutes a significant addition to our empirical assessment.

\subsection{Sythetic Network Type 1: 4 Node Random Network}
%In this section, we practically compare the loss functions on networks generated by algorithm \ref{alg:graph-generation}. We show how the models’ behaviors change as we alter the parameter p. Figure \ref{fig:type1-accuracy} shows models’ prediction accuracies for each case.

\begin{figure*}[!t]
    \centering
    \includegraphics[width=0.95\textwidth]{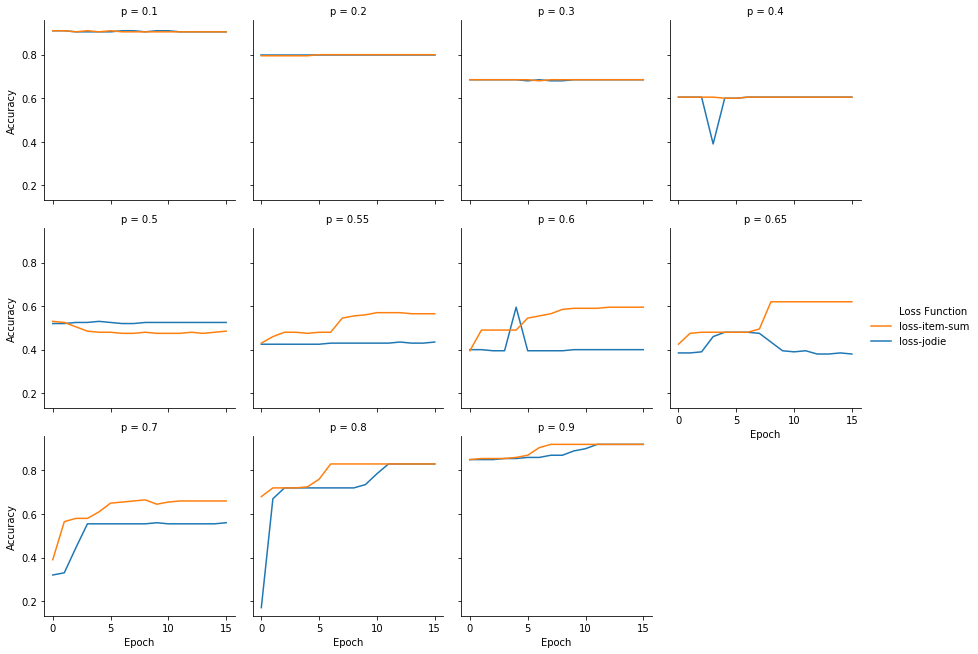}
    \caption{Comparison of edge prediction accuracy for user 3 edges in networks generated by Algorithm \ref{alg:graph-generation} using models with different loss functions.}
    \label{fig:type1-accuracy}
\end{figure*}

In this section, we empirically investigate the performance of the proposed loss functions on networks generated using Algorithm \ref{alg:graph-generation}. By manipulating the parameter $p$ in the generation process, we analyze how the models' behaviors change accordingly. Figure \ref{fig:type1-accuracy} illustrates the prediction accuracies of both models for each scenario.

Based on our theoretical analysis, we anticipate that the original loss function may yield incorrect predictions for $\frac{1}{2} < p < \frac{2}{3}$, and correct predictions outside of this range. In the plot, we observe that for $p \leq \frac{1}{2}$, both models exhibit nearly identical behavior. As expected, in the range $\frac{1}{2} < p < \frac{2}{3}$, the model utilizing the modified loss achieves the optimal prediction accuracy, while the original model makes erroneous predictions. For $p > \frac{2}{3}$, we expect both models to perform similarly. However, interestingly, for $p \in {0.8, 0.9}$, the $loss_{item-sum}$ converges to the optimum faster, possibly due to its more accurate loss function. The only observed discrepancy with our theoretical findings is in the case of $p = 0.7$. Despite expecting both functions to perform equally, $loss_{item-sum}$ still outperforms $loss_{tbatch}$ after ten epochs. This discrepancy may arise from the fact that ten epochs were insufficient for $loss_{tbatch}$ to reach its global minimum in this specific case.

%We expect from our theoretical analysis that the original loss function should make wrong predictions for $\frac{1}{2} < p < \frac{2}{3}$. In the plot, we see that for $p \leq \frac{1}{2}$ both models have behaved almost the same. The behavior in $\frac{1}{2} < p < \frac{2}{3}$ range is also as we expected. The model with modified loss could find the optimum prediction; however, the original model makes wrong predictions. For $\frac{2}{3} < p$, we expect that both models have the same behavior. Nevertheless, we see that for $p \in \{0.8, 0.9\}$, the $loss_{item-sum}$ reaches the optimum earlier, which may be because we have a more accurate loss function. The only observation that does not match our theoretical findings is for the case $p = 0.7$. In this case, we see a distance between the lines after ten epochs, which may be because ten epochs were not enough for $loss_{tbatch}$ to reach its global minimum.

\subsection{Synthetic Network Type 2: Repetitive Edge Network}
\begin{figure}[t]
    \centering
    \includegraphics[width=0.35\textwidth]{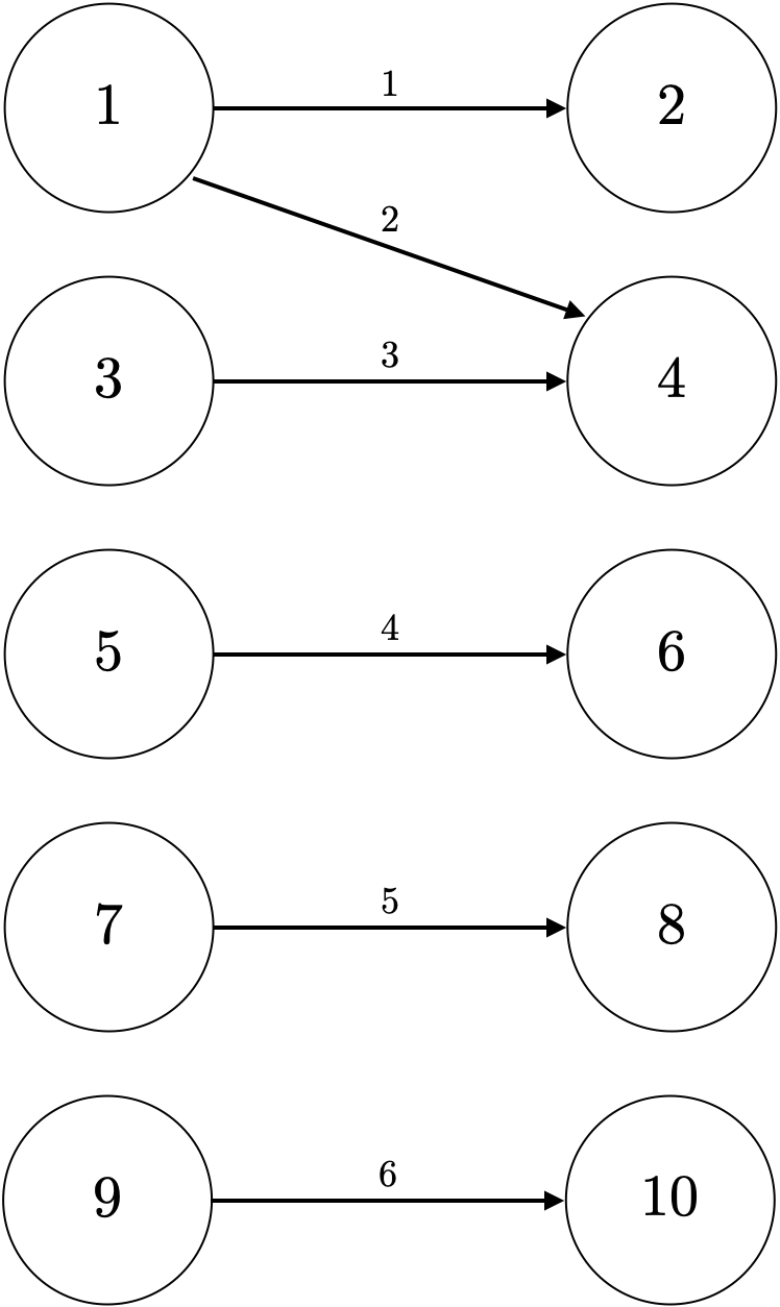}
    \caption{Illustration of edges in a network instance of type 2, featuring ten nodes and six edges. The presented set of interactions is subsequently repeated to generate additional interactions in the network.}
    \label{fig:type2-sample}
\end{figure}

%We now evaluate the link prediction task on a synthetic network structure with a repetitive deterministic set of edges. This structure has two consecutive edges for item one and then one edge for each user-item pair and repeats this set of edges for $k$ times. An example of base edges in a network of this type with ten nodes and six edges is illustrated in figure \ref{fig:type2-sample}.
In this section, we investigate a different network generation process that allows us to explore networks with a greater number of nodes. The synthetic interaction networks in this group are formed by repeating a deterministic set of edges, which can occur when user behaviors exhibit limited variation over time and follow a repetitive nature. For our analysis, we propose a simple network generation process characterized by deterministic repetitive interactions. The network generation process proceeds as follows: the first user establishes two consecutive edges with item one, followed by one edge for each user-item pair, and this set of edges is then repeated for a total of $k$ times. Figure \ref{fig:type2-sample} illustrates an example of the base edges in a network of this type with ten nodes and six edges.

%In these networks, predictions for users other than user 1 are straightforward. For user 1, the model should only learn that the edge connecting user 1 to item 2 comes before the edge with item 4. As this is a deterministic process, the global minimum of all loss functions happens when they could correctly predict both edges. However, because of the design of the network, (1, 2) edges end up in batches with a larger size relative to (1, 4) edges. Figure \ref{fig:type2-ratio} shows this for networks with different numbers of base edges.

\begin{figure}[!t]
    \centering
    \includegraphics[width=0.6\textwidth]{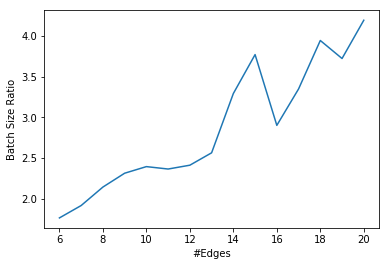}
    \caption{
The ratio of average batch sizes between edge (1, 2) and edge (1, 4) in type 2 networks is examined for various networks with different numbers of edges.}
    \label{fig:type2-ratio}
\end{figure}

In these networks, predictions for users other than user 1 are straightforward. However, for user 1, the model must learn to correctly predict that the edge connecting user 1 to item 2 precedes the edge connecting user 1 to item 4. Since this process is deterministic, all loss functions attain their global minimum when they can accurately predict both edges. However, due to the network's design, the edges (1, 2) end up in batches with larger sizes relative to the edges (1, 4). This disparity in batch sizes is depicted in Figure \ref{fig:type2-ratio}, which shows the ratio of average batch sizes of edges (1, 2) to edges (1, 4) in networks with varying numbers of edges.

%We investigate how this problem affects the prediction accuracy of models with each of the loss functions. As we could see in figure \ref{fig:type2-accuracy}, all of the models could eventually learn the pattern. Nevertheless, the number of epochs required for each of the methods varies. $loss_{tbatch}$ reaches the solution after $loss_{item-sum}$ and $loss_{full-sum}$ which depicts the problem with this loss function. Also, because in this problem, previous interaction histories are required for correct prediction for user 1, the model has to update its embeddings through time. As $loss_{full-sum}$ gives lower weights to regularization terms than $loss_{item-sum}$, it better allows the model to update embeddings and enables the model to reach the solution as early as the first epoch.

\begin{figure*}[!t]
    \centering
    \includegraphics[width=0.95\textwidth]{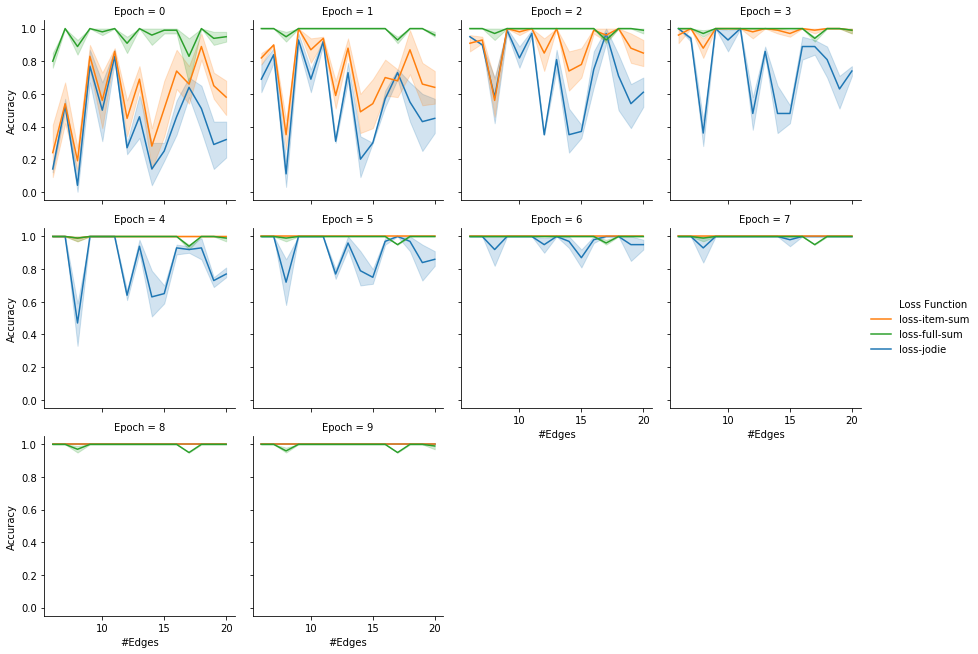}
    \caption{The accuracy of edge (1, 4) prediction for different loss functions in type 2 networks is evaluated through epochs.}
    \label{fig:type2-accuracy}
\end{figure*}

We proceed to investigate how this imbalance affects the prediction accuracy of models employing each of the loss functions. As shown in Figure \ref{fig:type2-accuracy}, all models ultimately learn the pattern. However, the number of epochs required for convergence varies across the methods. Specifically, $loss_{tbatch}$ reaches the solution after $loss_{item-sum}$ and $loss_{full-sum}$, indicating the presence of an issue with this loss function. Furthermore, since the correct prediction for user 1 necessitates knowledge of their previous interaction history, the model needs to update its embeddings over time. 
It is hypothesized that the lower weights attributed to regularization terms in $loss_{full-sum}$, as compared to $loss_{item-sum}$, enable the model to update embeddings more effectively, facilitating faster convergence towards the solution, potentially as early as the first epoch.

%Here, $loss_{full-sum}$ gives lower weights to regularization terms than $loss_{item-sum}$, we expect that these lower weights allow the model to update embeddings more effectively and reach the solution as early as the first epoch.

\subsection{Synthetic Network Type 3: Interaction Histories Sampled from Markov Model}
%network type 1 had a purely time-independent random construction, and network type 2 was completely deterministic. In designing network type 3, we use a random generation process such that, for having the best predictions, the model must use the users’ previous history. We construct each users’ interactions using a Markov model. Each user starts with the initial state, and by moving in the state space, we determine its edges. Figure \ref{fig:markov-model} describes the Markov model, and each node shows the item that the user will interact with if it goes in that state. So each user will have four edges sequentially with items in a path from the root node to a leave.

\begin{figure}[!t]
    \centering
    \includegraphics[width=0.5\textwidth]{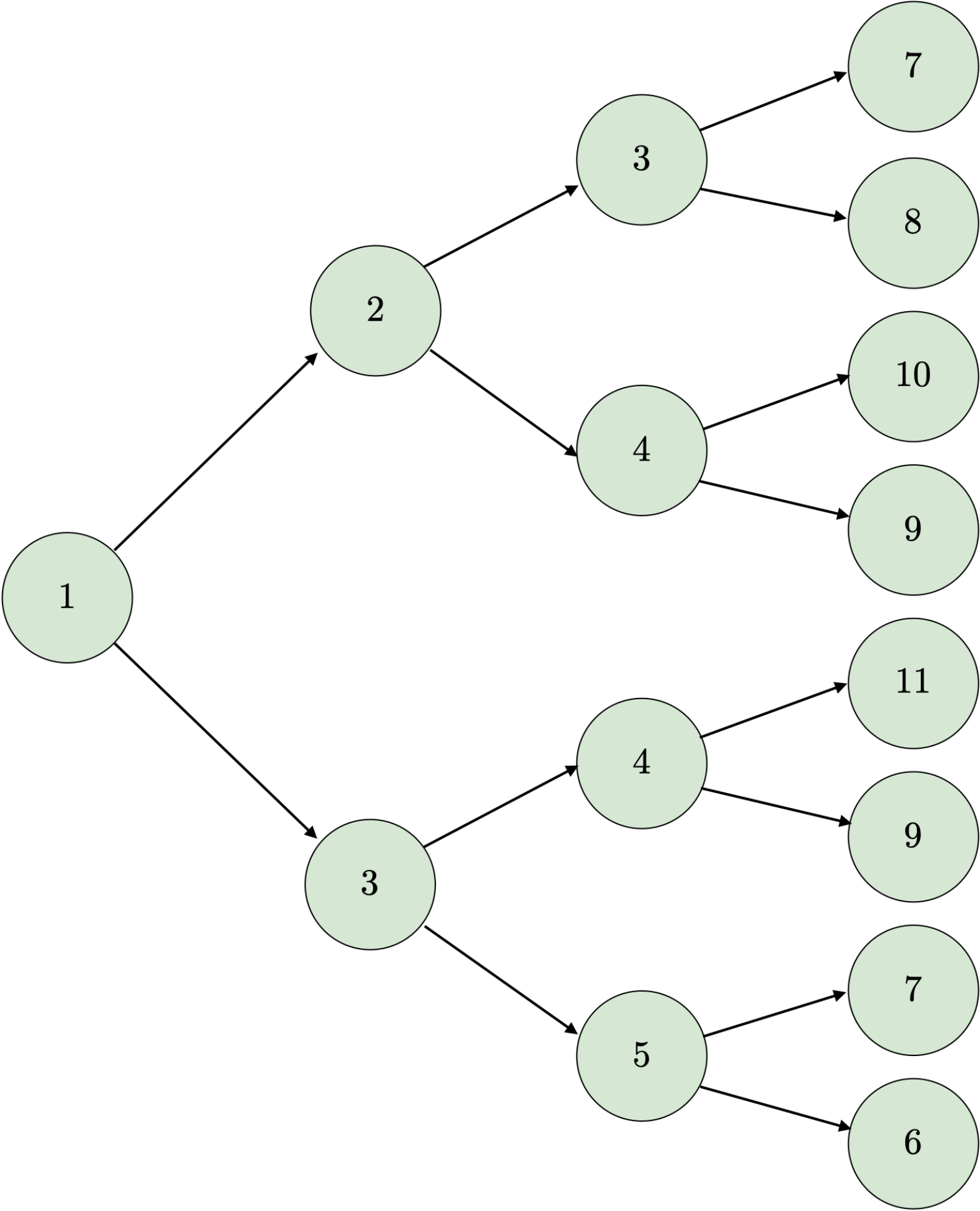}
    \caption{The Markov model used for generating network type 3}
    \label{fig:markov-model}
\end{figure}

In this network type, we aim to combine the random characteristic of network type 1 with the historical dependence observed in network type 2. To achieve this, we employ a random generation process that requires the model to utilize users' previous interaction history to achieve the best predictions. This process involves constructing each user's interactions using a Markov model, where each state represents the last item the user has interacted with. Users start with an initial state, and their subsequent interactions are determined by moving through the state space of the Markov model. Figure \ref{fig:markov-model} illustrates the Markov model utilized in this section, where each node corresponds to an item that the user will interact with if they traverse that particular state. Consequently, each user will have a sequence of four edges with items along a path from the root node to a leaf.

As a concrete example, let's consider a movie rating system with 11 movies. All users initially watch movie 1, and based on their viewing choice, they may proceed to watch either movie 2 or movie 3, with subsequent movie options depending on their previous choices.

%We evaluate the model based on the prediction accuracy of edges. Figure \ref{fig:type-3-accuracy} shows how loss functions’ behaviors vary. We could again see that $loss_{tbatch}$ has the lowest accuracy, while both $loss_{item-sum}$ and $loss_{full-sum}$ perform almost the same.

\begin{figure}[!t]
    \centering
    \includegraphics[width=0.5\textwidth]{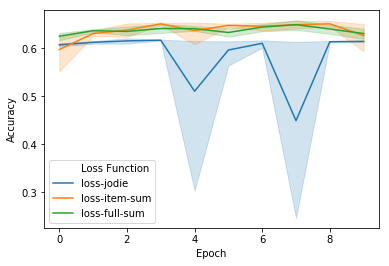}
    \caption{Prediction accuracy of different loss functions on a network instance of type 3.}
    \label{fig:type-3-accuracy}
\end{figure}

The model's evaluation is based on the prediction accuracy of edges in the synthetic networks generated using this Markov model. Figure \ref{fig:type-3-accuracy} illustrates the variations in the behaviors of the loss functions over epochs. Notably, $loss_{tbatch}$ exhibits the lowest accuracy, while both $loss_{item-sum}$ and $loss_{full-sum}$ perform comparatively better, with nearly identical results.

\subsection{Synthetic Network Type 4: Realistic Recommendation Scenario}

%In this scenario, we will focus on a more realistic network generation process. First, we construct a directed network over items, in which the $(i_s, i_d)$ link means that a user that had interacted with $i_s$ would consider $i_d$ as one of its next choices. So the outgoing neighbors of a node are the items that the user will interact with one of them in his next interaction. We also add some randomness in the process, so each user will select a completely random item with probability $p$, and with probability $1 - p$ it selects an item from the outgoing neighbors of the previous item that she has interacted with. The real-world situation that this model represents is a website that has some item suggestions on each item page, and the user may either select one of the suggestions or select another item randomly. Now to generate the whole interaction sequence, we will select users completely at random and sample the time difference between every two consecutive interactions of a user from an exponential distribution. We have also generated the directed graph over items using a k-out-degree model\cite{peterson2015distance} with preferential attachment, and $k = 10$. The other parameters used for the experiment are: $\#Users = 100, \#Items = 100, p = 1/4$.

\label{subsec:syn-net-4-recommendation}
In this scenario, we explore a more realistic network generation process that resembles recommendation systems commonly found in online platforms. These systems often present users with one or a few item suggestions for their next interaction. To simulate this, we construct a recommendation network as a directed network over items. Each directed edge $(i_s, i_d)$ indicates that a user who has interacted with item $i_s$ would consider item $i_d$ as one of their next choices. The outgoing neighbors of a node represent the items that a user may interact with in their subsequent interactions. We introduce some randomness in this process by allowing each user to select a completely random item with probability $p$, or with probability $1 - p$, the user selects an item from the outgoing neighbors of the previous item they interacted with. This setup reflects a real-world scenario where a website provides item suggestions on each item page, and users may either choose one of the suggestions or opt for another item randomly.

 The directed recommendation network over items is generated using a k-out-degree model~\cite{peterson2015distance} with preferential attachment and $k = 10$. The k-out-degree model ensures that all out-degrees are equal to $k$ and incorporates preferential attachment to mimic the idea that popular items are more likely to be recommended. After generating the recommendation network, we create the interaction sequence as follows: We randomly select users and sample the time difference between every two consecutive interactions from an exponential distribution. The parameters for the network generation process are: $|\mathcal{U}|= 100, |\mathcal{I}| = 100, p = 1/4$.

 We conducted experiments on ten samples generated from this process and reported the average results. 
 In our evaluation, we use two key metrics to assess the performance of the methods:
\begin{itemize}
    \item \textbf{MRR (Mean Reciprocal Rank)}: MRR is a statistical measure that assesses the quality of ranking methods. For a set of interactions $\mathcal{S}_{test}$ in the test set, the MRR is formally defined as follows:
\[MRR = \frac{1}{|\mathcal{S}_{test}|}\sum_{(u, j, t, f) \in \mathcal{S}_{test}}\frac{1}{rank_{j}}\]
Here, $rank_j$ represents the position of the true item that the user has interacted with in the model's recommendation list. A perfect MRR score of 1 indicates that the model correctly predicts the true item as the first recommendation for every interaction. 
    
    %Mean reciprocal rank is an statistic that evaluates ranking methods. For a set of $\mathcal{S}_{test}$ interactions in the test set, it is formally defined as $MRR = \frac{1}{|\mathcal{S}_{test}|}\sum_{(u, j, t, f) \in \mathcal{S}_{test}}\frac{1}{rank_{j}}$ where $rank_j$ is the rank of the true item that the user has interacted with, in our recommendation list. We will get $MRR = 1$ if the model predicts the true item for every interaction, as its first prediction.
    \item \textbf{R@10}: 
    R@10 measures the fraction of predictions in which the true item is included among the top ten recommendations made by the model. It quantifies the model's ability to capture relevant items within the first ten suggestions.
    %This is Recall @ 10, which indicates in which fraction of the predictions, the true item is within top ten recommendations of the model.
\end{itemize}
 Table \ref{table:syn4-results} presents the results for two cases with different numbers of interaction edges. The interaction dataset is split into training and test sets with a ratio of 16 to 1 for evaluation, and the size of the training set $|\mathcal{S}_{train}|$ is reported in the table. Notably, the proposed loss functions consistently outperformed the original loss function. Additionally, we observe that our proposed loss functions achieve higher percentage improvements and approach more accurate predictions, especially when trained with smaller datasets.

%We have tested the methods on 10 samples from the generation process and reported the average of the results. The results for two cases with different numbers of training edges are reported in Table \ref{table:syn4-results}. We see that our provided losses outperformed the original loss function, and also they got closer to the correct prediction with smaller training data. 

\begin{table}[th]
\caption{Impact of selecting different loss functions on various synthetic networks of type 4.}
\label{table:syn4-results}
\begin{center}
\begin{tabular}{|M{3cm}|M{2.2cm}|M{1.3cm}|M{1.3cm}|M{1.5cm}|M{1.5cm}|}
\hline
Dataset & Loss function & MRR & R@10 & MRR change\% & R@10 change\% \\
\hline
$\mathcal{S}_{train} = 8000$ & $loss_{tbatch}$ & 0.1967 & 0.5081 & - & -\\
$\mathcal{S}_{train} = 8000$ & $loss_{item-sum}$ & 0.2293 & \textbf{0.6584} & 16.57 & 29.58\\
$\mathcal{S}_{train} = 8000$ & $loss_{full-sum}$ &  \textbf{0.2301} & 0.6499 & 16.98 & 27.90 \\
\hline
$\mathcal{S}_{train} = 16000$ & $loss_{tbatch}$ & 0.2115 & 0.5648 & - & -\\
$\mathcal{S}_{train} = 16000$ & $loss_{item-sum}$ & \textbf{0.2286} & \textbf{0.6925} & 8.08 & 22.6\\
$\mathcal{S}_{train} = 16000$ & $loss_{full-sum}$ & 0.2276 & 0.6772 & 7.61 & 19.90\\
\hline
\end{tabular}
\end{center}
\end{table}

\subsection{Real Networks}

In this section, we conduct a comprehensive analysis of the performance of different loss functions on four distinct datasets. Additionally, we investigate how various characteristics of dynamic networks may influence the results.

The datasets used for evaluation are as follows:

\begin{itemize}
    \item \textbf{Reddit:} This dataset comprises interactions between users and subreddits within the Reddit platform. Each edge represents a post that a user has posted on a specific subreddit. The edge features consist of textual features collected from the posts.

    \item \textbf{Wikipedia:} The Wikipedia dataset contains edits made by users in the Wikipedia encyclopedia. The network represents users on one side and pages on the other side, with each edge corresponding to an edit performed on a page. Similar to the Reddit dataset, textual features of the edited text are available.
    \item \textbf{LastFM:} This dataset is based on the LastFM online music streaming platform. It represents the songs each user has listened to. Users have significant listening histories, while the number of unique songs each user has listened to is comparatively lower.
    \item \textbf{Myket:} Myket is an android application market that hosts over 350,000 applications for its users. We introduce this new dataset, which was collected by sampling a subset of users and recording their download interactions with frequently downloaded apps.
\end{itemize}

Table \ref{table:datasets} provides a description of the characteristics of each dataset. Notably, we observe variations in the number of interactions per user and the number of unique interactions for each user across different datasets. For instance, in the LastFM dataset, users exhibit a considerable listening history, while the number of unique songs each user has listened to is relatively lower. On the other hand, in the Wikipedia dataset, users have a lower number of interactions, and their interactions tend to be highly repetitive, with each user interacting with only two unique pages on average. The Myket dataset displays a different profile, with relatively lower repetition in interactions, primarily due to the fact that most users do not reinstall apps they have previously downloaded. 
%These dataset characteristics play a crucial role in evaluating the effectiveness of the loss functions on different dynamic networks.

\begin{table*}[th]
\caption{Charecteristics of real-world datasets used for evaluations.}
\label{table:datasets}
\begin{center}
\begin{tabular}{|M{1.8cm}|M{1.2cm}|M{1.2cm}|M{1.5cm}|M{2.5cm}|M{3cm}|}
\hline
Dataset & $|\mathcal{U}|$ & $|\mathcal{I}|$ & $|\mathcal{S}|$ & Average number of interactions for each user & Average number of unique items a user has interacted with\\
\hline
Myket & 10000 & 7988 & 694121 & 69.4 & 54.6 \\
LastFM & 980 & 1000 & 1293103 & 1319.5 & 158.2 \\
Reddit & 10000 & 984 & 672447 & 67.2 & 7.9 \\
Wikipedia & 8227 & 1000 & 157474 & 19.1 & 2.2 \\
\hline
\end{tabular}
\end{center}
\end{table*}

% Before getting to the evaluations, we will first study more characteristics of the networks and the characteristics of t-batching results on each of them. Figure \ref{fig:dist-tbatch} shows how the distribution of batch sizes varies between datasets. While the three original datasets follow almost the same pattern, the Myket dataset’s distribution has higher variance and a longer tail. This means that batches would have more heterogeneous sizes, which would cause higher bias in this dataset. Moreover, we analyze the datasets regarding how varied the transaction histories of each user are. We calculate the Average Entropy of users’ records and the percentage of interactions for each user’s most repeated item (Table \ref{table:datasets-stats}). We see that by both of the metrics, Myket and Reddit have a more diverse set of interactions for each user.

\begin{figure}[!t]
    \centering
    \begin{subfigure}[b]{0.48\textwidth}
		\centering
		\includegraphics[width=\textwidth]{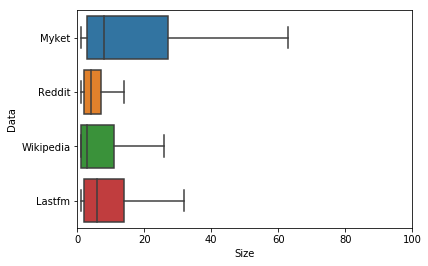}
		\label{fig:distribution-tbatch-violin}
	\end{subfigure}
	\begin{subfigure}[b]{0.48\textwidth}
		\centering
		\includegraphics[width=\textwidth]{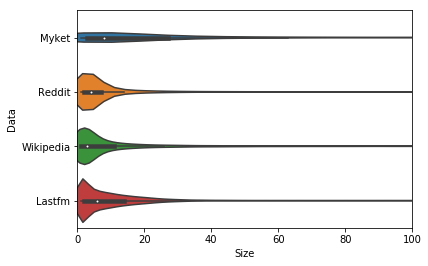}
		\label{fig:distribution-tbatch-violin}
	\end{subfigure}
    \caption{Distribution of training batch sizes for different datasets.}
    \label{fig:dist-tbatch}
\end{figure}

Prior to conducting the evaluations, we will conduct a more detailed examination of the networks' characteristics and the outcomes of t-batching on each dataset. Figure \ref{fig:dist-tbatch} depicts the distribution of batch sizes across datasets. While the three original datasets (Reddit, Wikipedia, and LastFM) exhibit similar patterns, the Myket dataset's distribution demonstrates higher variance and a longer tail. This indicates that batches in the Myket dataset would have more heterogeneous sizes, leading to a higher level of bias in this dataset. Additionally, we analyze the datasets in terms of the diversity of transaction histories for each user. We calculate the Average Entropy of users' records and the ratio of interactions corresponding to each user's most repeated item (Table \ref{table:datasets-stats}). The results reveal that both Myket and LastFM datasets showcase a more diverse set of interactions for each user, based on both metrics.

\begin{table}[th]
\caption{Network characteristics related to the diversity of user interaction records.}
\label{table:datasets-stats}
\begin{center}
\begin{tabular}{|M{3cm}|M{5.1cm}|M{3cm}|}
\hline
Dataset & Average ratio of interactions corresponding to the users' most
repeated item & Average Entropy \\
\hline
Myket & 0.095 & 3.718 \\
LastFM & 0.105 & 4.076\\
Reddit & 0.671 & 1.030 \\
Wikipedia & 0.865 & 0.304\\
\hline
\end{tabular}
\end{center}
\end{table}

% We will now present the evaluation metrics on each of the datasets and the loss functions (Table \ref{table:datasets-results}). We use the same evaluation metrics as Section \ref{subsec:syn-net-4-recommendation}. As we could see, the introduced loss functions have improvements on all of the datasets, indicating the studied problem of the original loss function also happens on real networks. We see that the effect is more assertive in the Myket dataset, where we see more than twenty percent gain by using altered loss functions. This behavior may be due to the more skewed distribution of batch sizes we have in the Myket dataset (Figure \ref{fig:dist-tbatch}). Additionally, we see that the $loss_{full-sum}$ has better results on datasets with lower interaction entropy. We believe this could be because, in these datasets, embeddings of users should change more rapidly. The loss function with reduced effect of regularization terms would allow these rapid changes and be beneficial for the modeling.

Table \ref{table:datasets-results} presents the evaluation metrics for each of the datasets using the different loss functions. The evaluation metrics are the same as those used in Section \ref{subsec:syn-net-4-recommendation}. As shown in the table, the proposed loss functions demonstrate improvements across all datasets, highlighting that the issue identified in the original loss function also affects real-world networks. Notably, the effect is more pronounced in the Myket dataset, with a gain of over twenty percent in the MRR metric achieved by using the altered loss functions. This behavior may be attributed to the more skewed distribution of batch sizes in the Myket dataset (Figure \ref{fig:dist-tbatch}). 
Moreover, in the case of the Reddit dataset, which exhibits low user interaction entropy, we note that the $loss_{item-sum}$ outperforms $loss_{full-sum}$. This observation may be attributed to the fact that $loss_{full-sum}$ places higher emphasis on the prediction error relative to the regularizer. However, in scenarios where users frequently interact with the same item, maintaining more static embeddings could prove advantageous, and thus, a higher level of regularization may be beneficial.
%Furthermore, it is observed that $loss_{full-sum}$ yields better results on datasets with lower interaction entropy. This suggests that in datasets where user embeddings need to adapt rapidly, a loss function with reduced impact of regularization terms can be advantageous for modeling purposes.

\begin{table}[th]
\caption{Impact of different loss functions on the evaluation datasets.}
\label{table:datasets-results}
\begin{center}
\begin{tabular}{|M{1.8cm}|M{2.2cm}|M{1cm}|M{1cm}|M{1.5cm}|M{1.5cm}|}
\hline
Dataset & Loss function & MRR & R@10 & MRR change\% & R@10 change\% \\
\hline
Myket & $loss_{tbatch}$ & 0.093 & 0.179 & - & -\\
Myket & $loss_{item-sum}$ & \textbf{0.120} & \textbf{0.208} & 29.38 & 16.29\\
Myket & $loss_{full-sum}$ & 0.118 & 0.200 & 26.94 & 11.85\\
\hline
LastFM & $loss_{tbatch}$ & 0.312 & 0.452 & - & -\\
LastFM & $loss_{item-sum}$ & 0.322 & 0.468 & 3.08 & 3.45\\
LastFM & $loss_{full-sum}$ & \textbf{0.323} & \textbf{0.471} & 3.45 & 4.23 \\
\hline
Reddit & $loss_{tbatch}$ & 0.724 & 0.845 & - & -\\
Reddit & $loss_{item-sum}$ & \textbf{0.737} & \textbf{0.885} & 1.80 & 4.88 \\
Reddit & $loss_{full-sum}$ & 0.681 & 0.747 & - & -\\
\hline
Wikipedia & $loss_{tbatch}$ & 0.768 & 0.831 & - & -\\
Wikipedia & $loss_{item-sum}$ & 0.760 & \textbf{0.847} & - & 1.91\\
Wikipedia & $loss_{full-sum}$ & \textbf{0.782} & 0.797 & 1.83 & -\\
\hline
\end{tabular}
\end{center}
\end{table}